\documentclass[journal]{IEEEtran}
\usepackage{tabularx} % in the preamble
\pdfoutput=1

\usepackage{hyperref}       % hyperlinks
\usepackage{url}            % simple URL typesetting
\usepackage{booktabs}       % professional-quality tables
\usepackage{amsfonts}       % blackboard math symbols
\usepackage{nicefrac}       % compact symbols for 1/2, etc.
\usepackage{microtype}      % microtypography
\usepackage{adjustbox}
\usepackage{hyperref}
\usepackage{bm}
\usepackage[caption=false]{subfig}

\usepackage[english]{babel}

\usepackage{multirow, rotating, array}
\usepackage{amsfonts, amssymb, amsmath}

\usepackage{floatrow}
\usepackage{flafter}

\usepackage{color, float}
\usepackage{afterpage}
\usepackage{flafter}

\usepackage{algorithmic}
\usepackage[algoruled]{algorithm2e}

\usepackage{wrapfig}
\usepackage{makeidx}
\usepackage{makecell}

\hypersetup{
	colorlinks   = true, %
	urlcolor     = blue, %
	linkcolor    = red, %
	citecolor    = blue   %
}

\newcommand{\add}[1]{\textcolor{magenta}{#1}}

\renewcommand{\add}[1]{#1}

\newcommand{\addminor}[1]{\textcolor{magenta}{#1}}

\renewcommand{\addminor}[1]{#1}

\IEEEpubid{\href{https://dx.doi.org/10.1109/TMI.2018.2883807}{\textbf{Published in IEEE Transactions on Medical Imaging, 2019}}}

\begin{document}

\title {A Recurrent CNN for Automatic Detection and Classification of Coronary Artery Plaque and Stenosis in Coronary CT Angiography}

%\title {A Recurrent Convolutional Neural Network for Automatic Detection and Classification of Coronary Artery Plaque and Stenosis in Coronary CT Angiography}

\author{
	Majd~Zreik, Robbert~W.~van~Hamersvelt, Jelmer~M.~Wolterink,\\Tim~Leiner, Max~A.~Viergever, Ivana~I\v{s}gum%
	\thanks{M.~Zreik, J.~M.~Wolterink, and I.~I\v{s}gum are with the Image Sciences Institute, University Medical Center Utrecht, The Netherlands (e-mail: m.zreik@umcutrecht.nl).}%
		\thanks{R.~W.~van~Hamersvelt is with the Department of Radiology, University Medical Center Utrecht, The Netherlands.}%
			\thanks{T.~Leiner is with the Department of Radiology, University Medical Center Utrecht and Utrecht University, The Netherlands.}
	\thanks{M. A.~Viergever is with the Image Sciences Institute, University Medical Center Utrecht and Utrecht University, The Netherlands.}
	\thanks{Copyright (c) 2018 IEEE. Personal use of this material is permitted. However, permission to use this material for any other purposes must be obtained from the IEEE by sending a request to pubs-permissions@ieee.org.}}

\markboth{}{}

\maketitle
	
%\footnote{corresponding author, \texttt{m.zreik@umcutrecht.nl}}

\begin{abstract}
	
Various types of atherosclerotic plaque and varying grades of stenosis could lead to different management of patients with coronary artery disease. Therefore, it is crucial to detect and classify the type of coronary artery plaque, as well as to detect and determine the degree of coronary artery stenosis.
This study includes retrospectively collected clinically obtained coronary CT angiography (CCTA) scans of 163 patients. In these, the centerlines of the coronary arteries were extracted and used to reconstruct multi-planar reformatted (MPR) images for the coronary arteries. To define the reference standard, the presence and the type of plaque in the coronary arteries (no plaque, non-calcified, mixed, calcified), as well as the presence and the anatomical significance of coronary stenosis (no stenosis, non-significant i.e. $<50\%$ luminal narrowing, significant i.e. $\geq 50\%$ luminal narrowing) were manually annotated in the MPR images by identifying the start- and end-points of the segment of the artery affected by the plaque.
To perform automatic analysis, a multi-task recurrent convolutional neural network is applied on coronary artery MPR images. First, a 3D convolutional neural network is utilized to extract features along the coronary artery. Subsequently, the extracted features are aggregated by a recurrent neural network that performs two simultaneous multi-class classification tasks. In the first task, the network detects and characterizes the type of the coronary artery plaque. In the second task, the network detects and determines the anatomical significance of the coronary artery stenosis.
The network was trained and tested using CCTA images of 98 and 65 patients, respectively. For detection and characterization of coronary plaque, the method achieved an accuracy of 0.77. For detection of stenosis and determination of its anatomical significance, the method achieved an accuracy of 0.80.
The results demonstrate that automatic detection and classification of coronary artery plaque and stenosis are feasible. This may enable automated triage of patients to those without coronary plaque and those with coronary plaque and stenosis in need for further cardiovascular workup.
\end{abstract}


%\linenumbers

\section{Introduction}

Coronary artery disease (CAD) is the most common type of heart disease \cite{AHA15}. CAD occurs when atherosclerotic plaque builds up in the wall of the coronary arteries. This may lead to stenosis, i.e. narrowing or occlusion of the coronary artery lumen, limiting blood supply to the myocardium and potentially leading to myocardial ischemia. Atherosclerotic plaque can be classified according to its composition into calcified plaque, non-calcified plaque, and mixed plaque, i.e plaque containing calcified and non-calcified components \cite{Raff2009}. Calcified plaque is considered stable and its amount in the coronary arteries is a strong predictor of cardiovascular events \cite{wolterink2016evaluation}. Unlike calcified plaque, non-calcified and mixed plaque are considered unstable and more prone to rupture. Such rupture may lead to acute coronary syndrome and could result in irreversible damage to the myocardium, i.e. myocardial infarction \cite{virmani2006pathology,Ache08}. As different types of plaque and varying grades of stenosis lead to different patient management strategies, it is crucial to detect and characterize coronary artery plaque and stenosis \cite{cassar2009chronic, cury2016cad}.

Coronary CT angiography (CCTA) is a well-established modality for identification, as well as for exclusion, of patients with suspected CAD. It allows for noninvasive detection and characterization of coronary artery plaque and grading of coronary artery stenosis \cite{Budo08a}. To day, these tasks are typically performed in the clinic by visual assessment \cite{cury2016cad}, or semi-automatically by first utilizing lumen and arterial wall segmentation and thereafter, defining the presence of plaque or stenosis \cite{Kiri13a}. However, the former suffers from substantial interobserver variability, even when performed by experienced experts \cite{Arba11}, while the latter is dependent on coronary artery lumen and wall segmentation which is typically time-consuming and cumbersome, especially in images with extensive atherosclerotic plaque or imaging artefacts \cite{Kiri13a}.
 	
Given the importance of plaque detection, a number of methods for coronary artery plaque detection and quantification have been proposed. Thus far, (semi-)automatic methods have been designed to detect either calcified or non-calcified plaque. Several methods have been developed to automatically segment and quantify calcifications in the coronary arteries in non-contrast CT and CCTA scans (e.g. \cite{wolterink2016evaluation,Wolt16,lessmann2017automatic,shadmi2018fully}). These methods employ machine learning to analyze axial reconstruction of CT scans. \IEEEpubidadjcol Typically, an excellent performance approaching the level of an expert is achieved \cite{wolterink2016evaluation}. On the contrary, automatic detection and quantification of non-calcified coronary plaque using CCTA has been less investigated. While the visual detection of calcified plaque is straightforward in CT/CCTA due to its higher CT density, the detection of non-calcified and especially mixed plaque is more challenging because of low contrast with adjacent tissues. Therefore, unlike methods detecting calcifications, standard visual evaluation as well as (semi-)automatic approaches detecting non-calcified plaque typically analyze straightened multi-planar reformatted (MPR) images. MPR images allow better visualization of the arterial lumen and identification of difficult to delineate non-calcified plaque in the arterial wall. To detect and quantify non-calcified plaque, previously proposed methods have performed manual or semi-automatic thresholding on CT values in predefined regions of interest \cite{schepis2009quantification,dey2010automated}. Typically, these methods require substantial manual interactions by experts. Even though the presence of mixed plaque is usually reported in clinical visual assessment, to the best of our knowledge, automatic methods detecting and quantifying such plaque have not yet been presented.

As stenosis detection and grading is highly important, a number of methods have been developed to (semi-)automatically detect and grade coronary artery stenosis in CCTA \cite{Kiri13a}. These methods either utilize machine learning approaches to analyze the vicinity of the coronary artery centerline (e.g. \cite{mittal2010fast,zuluaga2011automatic,sankaran2016hale}), or segment arterial lumen (e.g. \cite{halpern2011diagnosis,xu2012quantification,shahzad2012automatic,wang2012vessel,broersen2012frenchcoast}). 
Algorithms that utilize lumen segmentation for stenosis detection first delineate coronary artery lumen and subsequently detect and quantify stenosis by analyzing local changes and anomalies in the lumen of the delineated artery. Shahzad et al. \cite{shahzad2012automatic} first extracted the centerline of the artery and subsequently employed a graph cut approach and robust kernel regression to segment the arterial lumen. Thereafter, to detect and grade coronary stenosis, the diameter of the segmented lumen was compared with the expected diameter of a healthy lumen. The expected diameter of the healthy lumen was estimated by regression on the diameters of the lumina along the coronary artery. Wang et al. \cite{wang2012vessel} employed a level-set model to separately segment the inner and the outer arterial walls. Thereafter, a comparison between these arterial wall profiles enabled the detection and grading of stenosis.
Furthermore, algorithms that exploit machine learning for stenosis detection first compute a number of features along the centerline of an artery to describe local image intensities and arterial geometry. Subsequently, they use a supervised classifier to detect and quantify stenosis. For example, Zuluaga et al. \cite{zuluaga2011automatic} formulated arterial stenosis as anomalous vascular cross-sections along the artery centerline. The shape of the vascular cross-sections and their intensity profiles were described by hand-crafted features, and then the abnormal vascular cross-sections were detected by a support vector machine. Sankaran et al. \cite{sankaran2016hale} first estimated healthy diameters of the coronary arteries using downstream and upstream properties of coronary tree vasculature as features for random forest regressors. Then, the degree of a stenosis was estimated based on the ratio of the local artery diameter, estimated using maximum inscribed spheres, to the estimated healthy diameter.

\add{Here, we present a method to automatically detect and characterize the type of coronary artery plaque, as well as to detect and determine the anatomical significance of coronary artery stenosis in CCTA scans. 
To perform the automatic comprehensive analysis, a multi-task recurrent convolutional neural network (RCNN) is employed to analyze the vicinity along the extracted centerline in an MPR image and to simultaneously carry out two classification tasks. In the first task, the network detects and characterizes the type of the coronary artery plaque, i.e. no plaque, non-calcified, mixed or calcified plaque. In the second task, the network detects and determines the anatomical significance of the coronary artery stenosis, i.e. no stenosis, non-significant or significant stenosis. The RCNN analyzes the vicinity along the artery centerline, which is defined as a sequence of small volumes along the centerline. This definition enables the RCNN, built from a 3D convolutional neural network (CNN) and a recurrent neural network (RNN) connected in series \cite{Lecu15}, to extract image features from smaller volumes regardless of plaque length, and then to aggregate all features extracted along the plaque. 
Our contributions are fourfold. Firstly, we propose to jointly classify plaque and stenosis, while previously proposed methods have detected either plaque or stenosis. Secondly, unlike previous automatic methods, our method does not require segmentation of the coronary artery lumen and/or wall nor exploiting geometric information about the artery lumen. Instead, it only requires the coronary artery centerline. Thirdly, we are the first to use deep neural networks to approach the task of coronary artery plaque and stenosis analysis, or more specifically a \textit{3D-CNN} paired with an RNN to analyze medical data. Finally, previous works for classifying coronary artery plaque require detailed reference annotations for each voxel affected by the plaque. This kind of manual annotations is extremely challenging and time consuming. In the presented work, we employ only weakly annotated reference (start- and end-points of a lesion with a single label for all voxels in that lesion) to detect and characterize the plaque and thereby substantially simplify the manual annotation procedure.}

The remainder of the manuscript is organized as follows. Section \ref{data} describes the data and reference standard. Section \ref{method} describes the method and Section \ref{eval} describes the evaluation. Section \ref{results} reports our experimental results, which are then discussed in Section \ref{discussion}.

\section{Data} \label{data}
	
\subsection{Patient and image data}
	
This study includes retrospectively collected clinically obtained CCTA scans of 163 patients (age: $59.2 \pm 8.8$ years, 126 males) acquired in our hospital between 2012 and 2016. The Institutional Ethical Review Board waived the need for informed consent. 
%All CCTA scans were acquired using an ECG-triggered sequential protocol or retrospectively ECG-gated spiral protocol on a 256-slice scanner (Philips Brilliance iCT, Philips Medical, Best, The Netherlands). A weight dependent tube voltage and tube current were used; for sequential 100-120 kVp, 210-300 mAs and for spiral 100-120 kVp, 600-700 mAs. Contrast medium was injected using a weight dependent protocol with a flow rate of 6-6.7 mL/s for a total of 70-80 mL iopromide (Ultravist 300 mg I/mL, Bayer Healthcare, Berlin, Germany), followed by a 50-67 mL mixed contrast medium and saline (50:50) flush, and a 30-40 mL saline flush. Images were reconstructed to an in-plane resolution ranging from 0.38 to 0.56 mm, with 0.9 mm slice thickness and 0.45 mm slice increment
All CCTA scans were acquired using an ECG-triggered step-and-shoot protocol on a 256-detector row scanner (Philips Brilliance iCT, Philips Medical, Best, The Netherlands). A tube voltage of 120 kVp and tube current between 210 and 300 mAs were used. For patients $\le80$ kg contrast medium was injected using a flow rate of 6 mL/s for a total of 70 mL iopromide (Ultravist 300 mg I/mL, Bayer Healthcare, Berlin, Germany), followed by a 50 mL mixed contrast medium and saline (50:50) flush, and a 30 mL saline flush. For patients $>80$ kg the flow rate was 6.7 mL/s and the volumes of the boluses were 80, 67 and 40 mL, respectively. Images were reconstructed to an in-plane resolution ranging from 0.38 to 0.56 mm, with 0.9 mm slice thickness and 0.45 mm slice increment. 

In each CCTA image, centerlines of the coronary arteries were extracted using the method previously described by Wolterink et al. \cite{wolterink2018coronary}. The method requires manual placement of a single seed point in the artery of interest, after which the arterial centerline is extracted between the ostium and the most distal point as visualized in the CCTA image. Using the extracted centerlines, a 0.3 mm isotropic straightened MPR image was reconstructed for each artery and used for further analysis. 

\subsection{Reference standard}\label{ref_stand}

\begin{figure}[h]
		\includegraphics[width=1\linewidth]{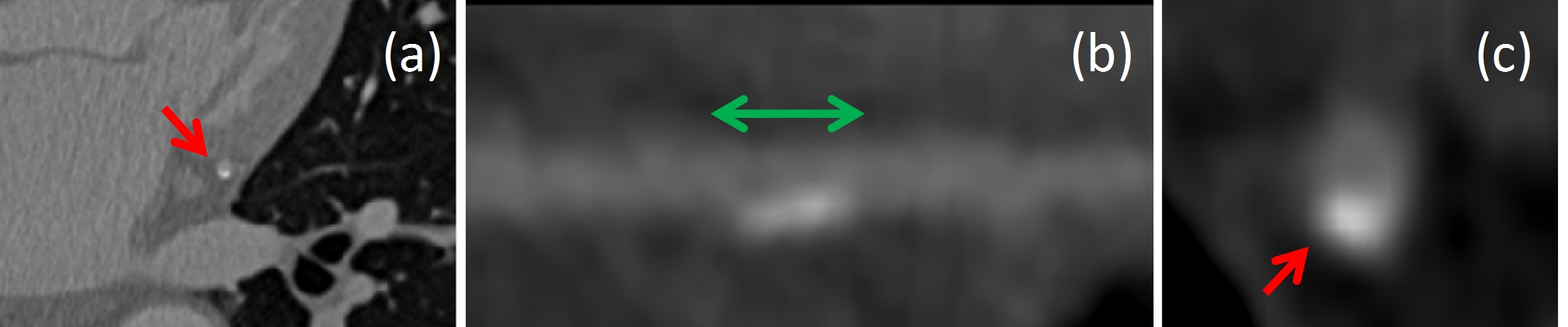}
		\caption{ (a) Axial and (b) straightened MPR image with longitudinal view, and (c) cross-sectional view showing coronary artery in CCTA. This artery contains a lesion (spanning the green arrow) labeled as containing calcified plaque with non-significant stenosis. Red arrows indicate the location of this plaque in other views.}
		\label{fig:example}
		
	\end{figure}

To define a reference standard for atherosclerotic plaque and coronary artery stenosis, MPR images of coronary arteries were employed (Fig.~\ref{fig:example}). As only arteries with a diameter greater than 1.5 mm are clinically evaluated with CCTA \cite{cury2016cad}, only those were annotated in this study. Plaque type and anatomical significance of the stenosis were manually annotated by an expert using custom-built software and following the guidelines of the Society of Cardiovascular Computed Tomography (SCCT) for reporting coronary artery disease \cite{cury2016cad}. For each plaque, the expert marked its start- and end-points, its type (\textit{non-calcified}, \textit{calcified}, or \textit{mixed} i.e. containing both non-calcified and calcified components), and the anatomical significance of the stenosis caused by the plaque (\textit{non-significant} i.e. with $<50\%$ luminal narrowing, \textit{significant} i.e. with $\geq 50\%$ luminal narrowing). The significance of a stenosis was determined by visual estimation of the maximal grade of luminal narrowing caused by the plaque. Note that the plaque voxels were not segmented, but only the part of the artery containing the plaque was identified (Fig.~\ref{fig:example}b). The expert also annotated a number of segments of arteries without plaque and stenosis in different patients. We refer to the annotated and analyzed parts of the arteries as segments. However, note that these are not the anatomically defined coronary artery segments.

%As patient management and treatment strategies depend on diagnosis at the segment-, artery- and patient-level \cite{cassar2009chronic,cury2016cad}, we also evaluated the ability of the proposed method to detect stenosis and to classify its anatomical significance on an artery- and patient-level. Therefore, an artery was defined as containing significant stenosis if it contained at least one segment with significant stenosis. Similarly, an artery was defined as containing a non-significant stenosis if it contained at least one segment with non-significant stenosis, but it did not contain any segment with significant stenosis. An artery was defined as having no stenosis if none of its segments contained stenosis. In a like manner, a patient was defined as containing significant stenosis if at least one of its arteries contained a significant stenosis. Similarly, a patient was defined as containing non-significant stenosis if at least one of its arteries contained a non-significant stenosis and no artery contained a significant stenosis. A patient was defined as having no stenosis if none of its arteries contained stenosis.

As patient management and treatment strategies depend on diagnosis at the segment-, artery- and patient-level \cite{cassar2009chronic,cury2016cad}, we additionally evaluate the ability of the proposed method to detect stenosis and to classify its anatomical significance on an artery- and patient-level. Therefore, segment-level labels provided by the expert were translated to the artery- and patient-level as follows. Arteries were labeled according to the most severe stenosis significance among their segments. If no stenosis of any significance was found in any of the annotated segments, the artery was considered non-stenotic. Likewise, patients were labeled according to the most severe stenosis significance among their arteries, or considered as having no stenosis.

The dataset of 163 patients contained 1259 manually labeled arterial segments in 534 arteries. The manually annotated segments included 37 non-calcified, 161 mixed and 317 calcified plaques that caused non-significant stenosis. Additionally, there were segments with 29 non-calcified, 91 mixed and 41 calcified plaques that caused significant stenosis. Moreover, 583 segments without plaque and stenosis were annotated. Of the annotated segments, 528 were in the left anterior descending artery (LAD) or one of its branches, 305 were in the left circumflex artery (LCX) or one of its branches, and 426 were in the right coronary artery (RCA) or one of its branches.

To assess the interobserver agreement, a second trained observer, blinded to the reference standard, annotated the same set of arteries following the same guidelines.
%\textcolor{red}{In cases that a region is non gradable due to poor image quality, artifacts or motion or due to small artery diameter ($<2 mm$), that region was labeled as non-gradable, as shown in Table~\ref{list:stenosis_grading}. }

	\section{Method} \label{method}

\begin{figure}[h]
	\includegraphics[width=0.9\linewidth]{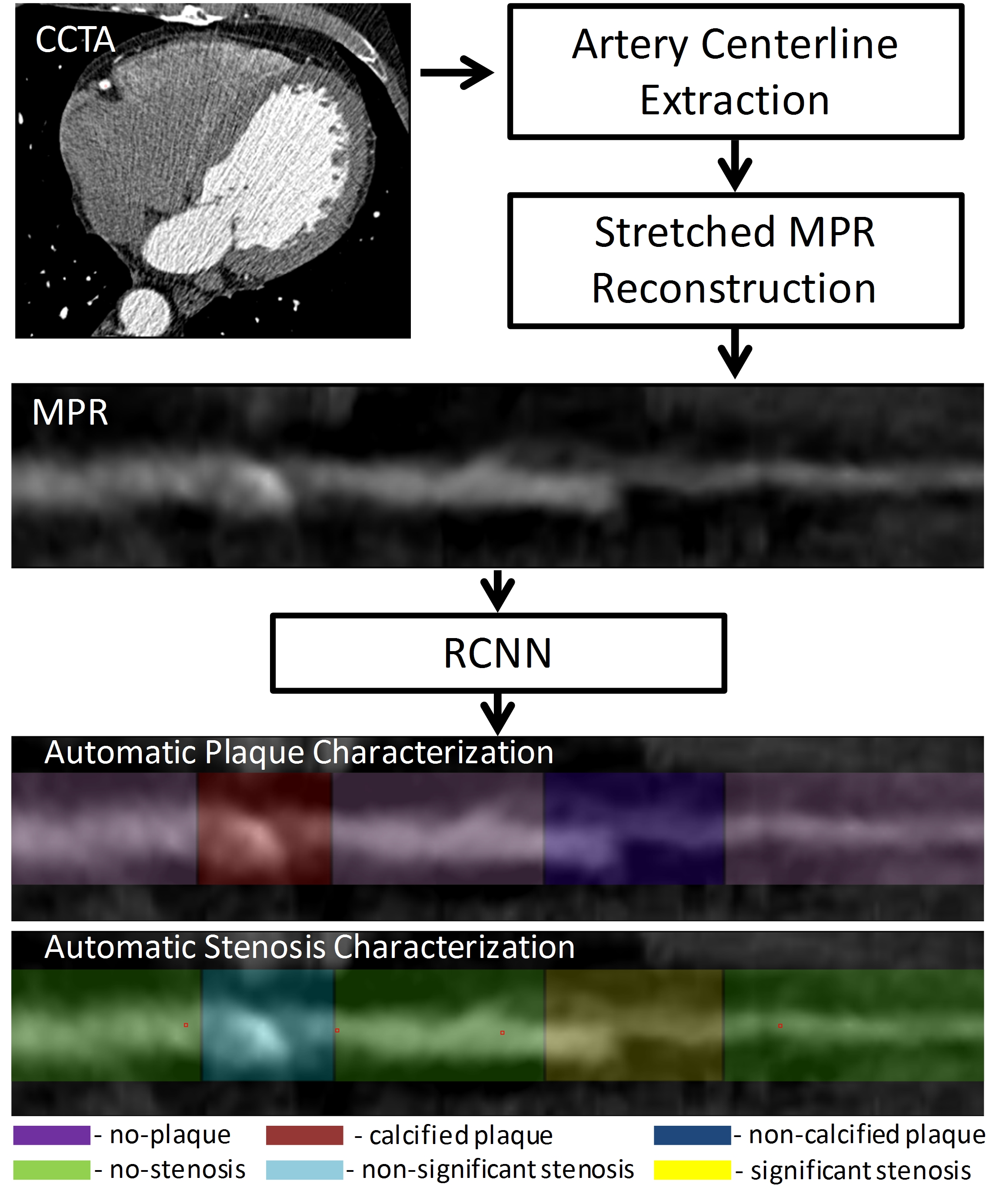}
	\caption{\add{Illustration of the proposed workflow. In a CCTA scan, the centerlines of the coronary arteries are extracted and used to reconstruct stretched multi-planar reformatted (MPR) images for the coronary arteries. To perform the automatic analysis, a multi-task recurrent convolutional neural network (RCNN) is applied to coronary artery MPR images to perform two simultaneous multi-class classification tasks. In the first task, the network detects and characterizes the type of the coronary artery plaque (no plaque, non-calcified, mixed, calcified). In the second task, the network detects and determines the anatomical significance of the coronary artery stenosis (no stenosis, non-significant i.e. $<50\%$ luminal narrowing, significant i.e. $\geq 50\%$ luminal narrowing)}.
}
	\label{fig:flow}
	
\end{figure}

	To detect and characterize the type of coronary artery plaque, as well as to detect and determine the anatomical significance of coronary artery stenosis, an RCNN is designed.  \add{An illustration of the proposed workflow is shown in Fig.~\ref{fig:flow}}.
	Recently, RCNNs have been successfully used for video recognition and description (e.g. \cite{donahue2015long,karpathy2015deep,ng2015beyond}), object recognition (e.g. \cite{liang2015recurrent}), speech modeling \cite{de2015survey}, and in medical image analysis (e.g. \cite{poudel2016recurrent,Litj17,andermatt2016multi,xue2018full}). 
	RCNNs connect a CNN with an RNN in series to analyze a sequential input. The CNN extracts image features for each element of the sequence independently (e.g. frame in a video clip, word in a sentence, cardiac phase in cardiac cycle), and these extracted features are then fed to the RNN that analyzes the relevant sequential dependencies in the whole sequence.

	In this work, the reference was defined following clinical practice where parts of the arteries are classified with respect to plaque and stenosis. Hence, only arterial segments containing plaque were identified, instead of e.g. annotating all cross-sections of the arterial lumen or all voxels in the arterial wall.
	 Given that the appearance of the plaque along the whole segment is important for characterization of its type and for determination of stenosis presence and significance, the artery along the entire segment needs to be analyzed. Instead of extracting a single, possibly large, volume covering the segment at hand, we represent the segment as a sequence of small volumes along its centerline. This enables us to employ a relatively shallow CNN to extract image features from smaller volumes independently. A shallow CNN that analyzes smaller volumes may have fewer parameters and is therefore less prone to overfitting. To aggregate and analyze the features extracted from all small volumes along the segment, we employ an RNN. For the whole analyzed sequence, the network 1) detects and characterizes coronary artery plaque  
	 i.e. classifies the segment as either containing no plaque, containing non-calcified, mixed or calcified plaque, and 2) detects and determines the anatomical significance of coronary artery stenosis, i.e. classifies the segment as either containing no stenosis, or containing non-significant or significant stenosis.

\begin{figure}[h!]

		\includegraphics[width=0.9\linewidth]{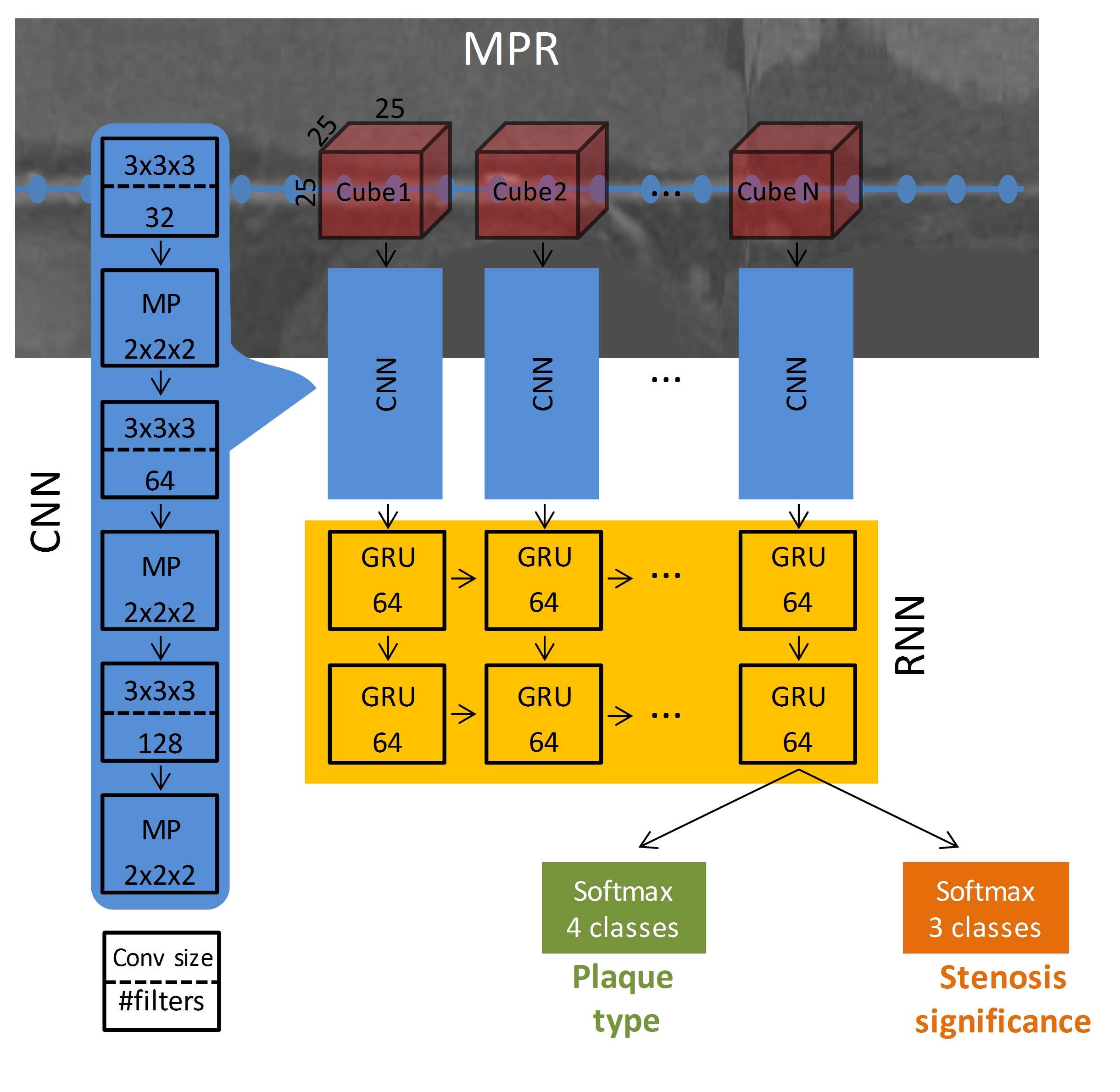}
		
		\caption{Overview of the proposed network. An MPR is obtained using the artery centerline points (blue dotted line). The input of the network is a sequence of cubes extracted from the MPR, along the artery centerline. A CNN extracts features out of $25\times25\times25$ voxels cubes. Subsequently, an RNN processes the entire sequence using gated recurrent units (GRUs). The output of the RNN is fed into two softmax classifiers to simultaneously characterize plaque and stenosis.}
		\label{fig:cnn_rnn}
	\end{figure}

	 An illustration of the proposed RCNN is shown in Fig.~\ref{fig:cnn_rnn}. The input of the network is a sequence with a maximum length of 25 cubes of $25\times25\times25$ voxels with stride of 5 voxels extracted from the MPR along the coronary artery centerline. The maximal length was chosen based on the longest annotated plaque in the training set. The size of the cube was chosen so that it contains the whole arterial lumen and the vicinity of the artery that may be needed in case of positive remodeling \cite{cury2016cad}. Each cube is analyzed by a 3D CNN. The CNN consists of three convolutional layers with kernels of $3\times3\times3$ elements, with 32, 64, 128 filters, respectively. Each convolutional layer is followed by a $2\times2\times2$ max-pooling layer and batch normalization \cite{Ioff15}.
	The features extracted by the CNN are fed to the RNN. The RNN consists of 2 layers of 64 Gated Recurrent Units (GRUs) \cite{cho2014learning} each. Rectified linear units (ReLU) \cite{Glor11a} are used in both CNN and RNN layers as activation functions, except for the output layer of the RNN. To perform both classification tasks simultaneously, the output of the last layer of the RNN is fed into two separate multi-class softmax classifiers. The first classifier has four output units for detection of plaque and characterization of its type (no-plaque, non-calcified, mixed, calcified). The second classifier has three output units for detection of stenosis and determination of its anatomical significance (no-stenosis, non-significant stenosis, significant stenosis). The RCNN has a total number of 340,295 parameters.

\section{Evaluation}\label{eval}

Performance of the network was evaluated on segment-, artery- and patient-levels. For segment-level evaluation, only the predicted labels along the centerlines that fall within the manually annotated segment boundaries were considered. For artery-level evaluation, all predicted labels along the complete artery centerline were taken into account. For patient-level evaluation, all predicted labels along the complete centerlines of all arteries of a patient were taken into account. 

An automatically labeled segment is considered to be a true positive in the classification of plaque type or stenosis significance when it has an overlap of at least 1 mm with a manually annotated segment sharing its label. On the contrary, a segment is considered to be true positive in the detection of plaque absence (\textit{no plaque}) or stenosis absence (\textit{no stenosis}), only when no point along the segment has any plaque or stenosis, respectively.

As most patients have multiple arterial plaques of different types, the evaluation of plaque detection and characterization was performed on a segment-level only. The average accuracy of the prediction over all segments and labels, i.e. the average percentage of correctly labeled segments, was computed. To assess the overall performance, the unweighted average of F1 score was computed. This was done for each label separately, and then computing the unweighted mean across all labels, averaged over all segments. Given the multiple categories of the plaque labels, unweighted Cohen’s $\kappa$ metric was used to measure the reliability between the predicted plaque labels and the reference standard.
 
The evaluation of stenosis detection and characterization was performed on segment-, artery- and patient-levels.
Automatically determined stenosis significance for an artery or a patient is considered true positive if any of the automatically detected labels along the artery centerline or patients arteries match the reference label of that artery or patient, respectively. %Similarly, automatically determined stenosis significance for a patient is considered true positive if the automatically detected labels of any artery match the reference label of that patient.
On the contrary, an artery or patient is considered to be true positive in the detection of stenosis absence (\textit{no stenosis}) only when no stenosis was detected at any point along the artery or in any of the patient's arteries, respectively.
As for plaque evaluation, the average accuracy and the unweighted average F1 score were computed to assess the overall agreement for predicting the stenosis labels. Given the grading of the stenosis, Cohen’s linearly weighted $\kappa$ metric was used to measure the reliability between the predicted stenosis labels and the reference standard.
 
Interobserver reliability for both plaque and stenosis analyses was assessed using the same metrics by comparing the annotations of the second observer to the reference standard.

%To evaluate the performance on segment- and artery-levels, labels of points along the coronary artery centerline need to be converted into single label for a segment or  an artery. Therefore, for segment-based evaluation, the entire segment if the method predicts a matching label to any of the centerline points within the manually annotated segment boundaries. For artery-level evaluation, a label is assigned if the automatic method predicted the corresponding label to any of the centerline points within the entire artery. 

\section{Experiments and results}\label{results}
From the available dataset consisting of 163 patients, CCTA scans of 81 (50\%) and 17 (10\%) patients were randomly chosen for training and validation, respectively. The scans of the remaining 65 (40\%) patients were used for testing the method. All CNN and RNN hyperparameters were determined in preliminary experiments using the training and validation scans only.

Prior to training, several data augmentation techniques were utilized to increase the training set size. First, to make the network invariant to rotations around the artery centerline, random rotations between 0 and 360 degrees around the coronary artery centerline were applied to all cubes of a sequence. Second, to make the network invariant to slight inaccuracies in manual annotations of the points defining the segment, a sequence of a segment was varied by randomly choosing centers of cubes with a stride of 5 voxels with a uniform random shift between $\pm3$ voxels along the MPR centerline. Third, to make the network robust to possible inaccuracies in the extraction of the coronary artery centerline, the center of each cube was randomly shifted around its origin by up to 2 voxels, in any direction.

The network was trained with mini-batches containing only the manually annotated segments. In the dataset, the distribution of plaque types and stenosis grades is unbalanced (Section \ref{ref_stand}). To avoid a potential bias towards the most common type of plaque and stenosis in the dataset (i.e. calcified plaque with non-significant stenosis), a stratified random data sampling was performed during the training. Each training iteration included two distinct mini-batches. One mini-batch contained segments balanced with respect to their plaque classes regardless of the stenosis significance. A second mini-batch contained segments balanced with respect to the stenosis classes regardless of the plaque type. For each segment, a sequence of cubes, spanning the entire length of the segment, was extracted from the MPR volume along the artery centerline. 
The categorical cross-entropy was employed as loss function of each softmax classifier \add{and the L2 regularization was used with $\gamma=0.001$ for all layers.} The loss of the network was defined as the average of the two individual losses \add{(Eq.~\ref{loss})}. %R1C1 

\add{
\begin{equation}\label{loss}
L=-\frac{1}{2}(\sum_{i} y^{p}_{i}log(p^{p}_{i})+\sum_{j} y^{s}_{j}log(p^{s}_{j})) + \frac{\gamma}{2}\sum_{k} w_{k}^2
\end{equation}
where $y^{p}$ and $y^{s}$ are one-hot encoded vectors for the labels of plaque ($i=0,1,2,3$) and stenosis ($j=0,1,2$), respectively, and $p^{p}$ and $p^{s}$ are the softmax output probabilities for the plaque and stenosis, respectively. $w_k$ is a trainable weight in the network ($k=0,...,N_{param}$), where $N_{param}$ is the total number of trainable parameters in the network. }
%R1C2

\begin{figure}[h!]
	\centering
	\subfloat[]{\includegraphics[width=1\linewidth]{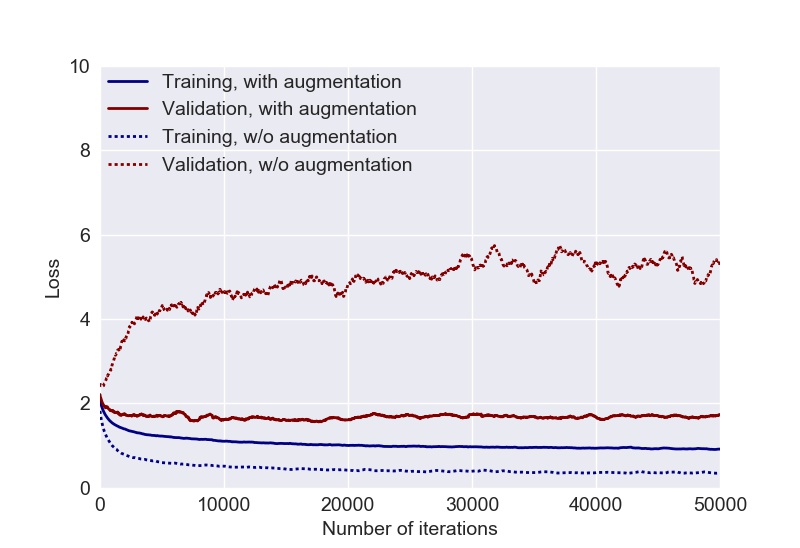}}
		
	\hfil
	
	\subfloat[]{\includegraphics[width=1\linewidth]{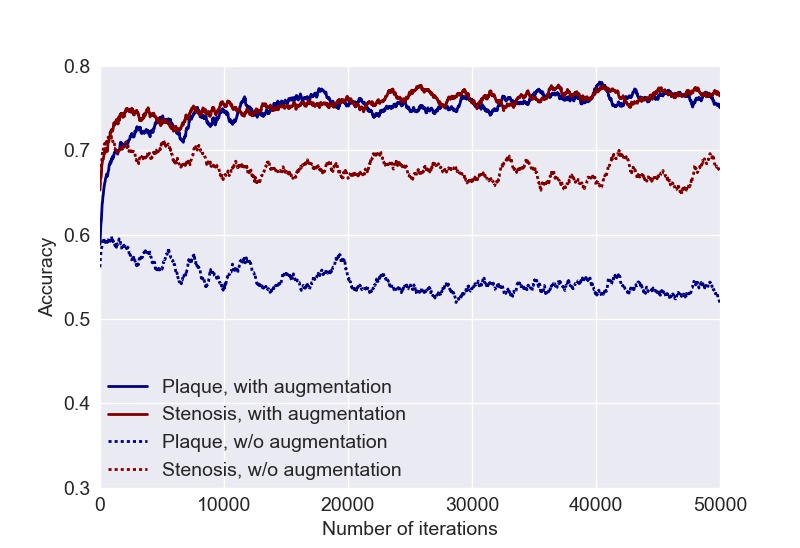}}
	\caption{(a) Training and validation losses during training with and without data augmentation. (b) Validation accuracies for detection and characterization of coronary artery plaque, as well as detection and determination of the anatomical significance of coronary artery stenosis during training with and without data augmentation.}
	\label{learning_curve}
\end{figure}

\add{During training,} mini-batches of 36 sequences \add{(i.e. segments)} were used to minimize the loss function with Adam optimizer \cite{kingma2014adam} with \add{constant} learning rate $0.001$, and a random dropout \cite{Sriv14} of 50\% was applied in each recurrent layer to prevent overfitting. \add{Training was performed for 50,000 iterations (Fig.~\ref{learning_curve}(a)). To evaluate the effect of data augmentation during training, an identical network was trained without any augmentation and the accuracies during training with and without data augmentation are illustrated in Fig.~\ref{learning_curve}(b). In these figures, the stability and the convergence of the training process with data augmentation are shown, while training without data augmentation shows signs of overfitting on the training set where accuracies on the validation set decrease.}

Unlike in the training phase, where the defined start- and end-points of a segment were used, during testing, segments were not defined, and labels were predicted along the whole analyzed artery. Therefore, all points along the coronary centerline were classified and labeled by the network. This was performed by feeding the network a fixed-length sequence centered around each centerline point. The fixed-length sequence consisted of 5 cubes with a stride of 5 voxels extracted along the coronary artery centerline. These parameters were optimized in preliminary experiments on the training set. The output probabilities for plaque and stenosis were then assigned to the center of the evaluated sequence. Thereafter, a label for plaque and a label for stenosis was defined for each centerline point by the class having the highest probability in each task separately.

 \begin{figure}[h!]
	\centering
	\subfloat[]{\includegraphics[width=1\linewidth]{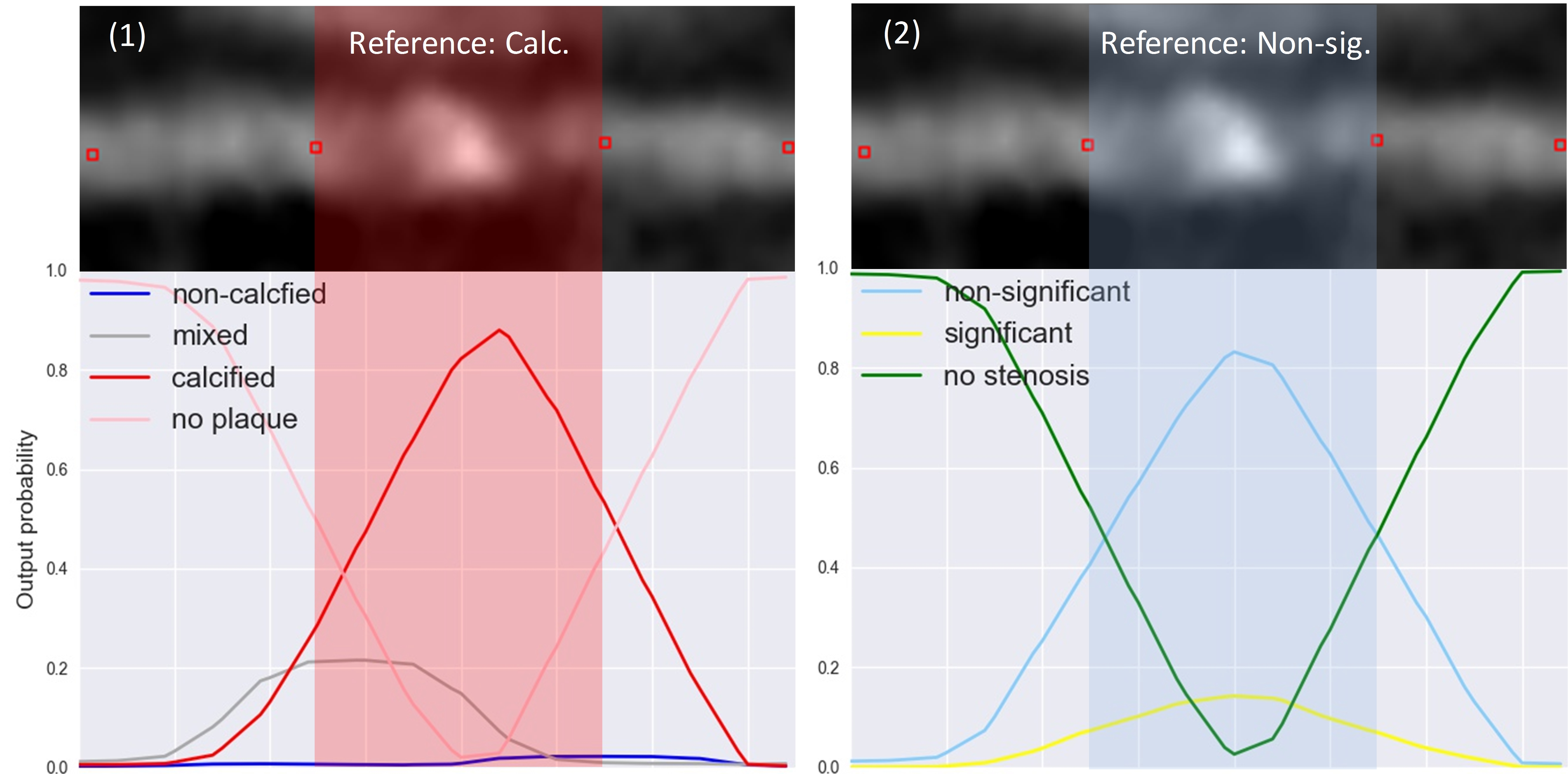}}
		\label{fig:prob_good}
	\hfil
	
	\subfloat[]{\includegraphics[width=1\linewidth]{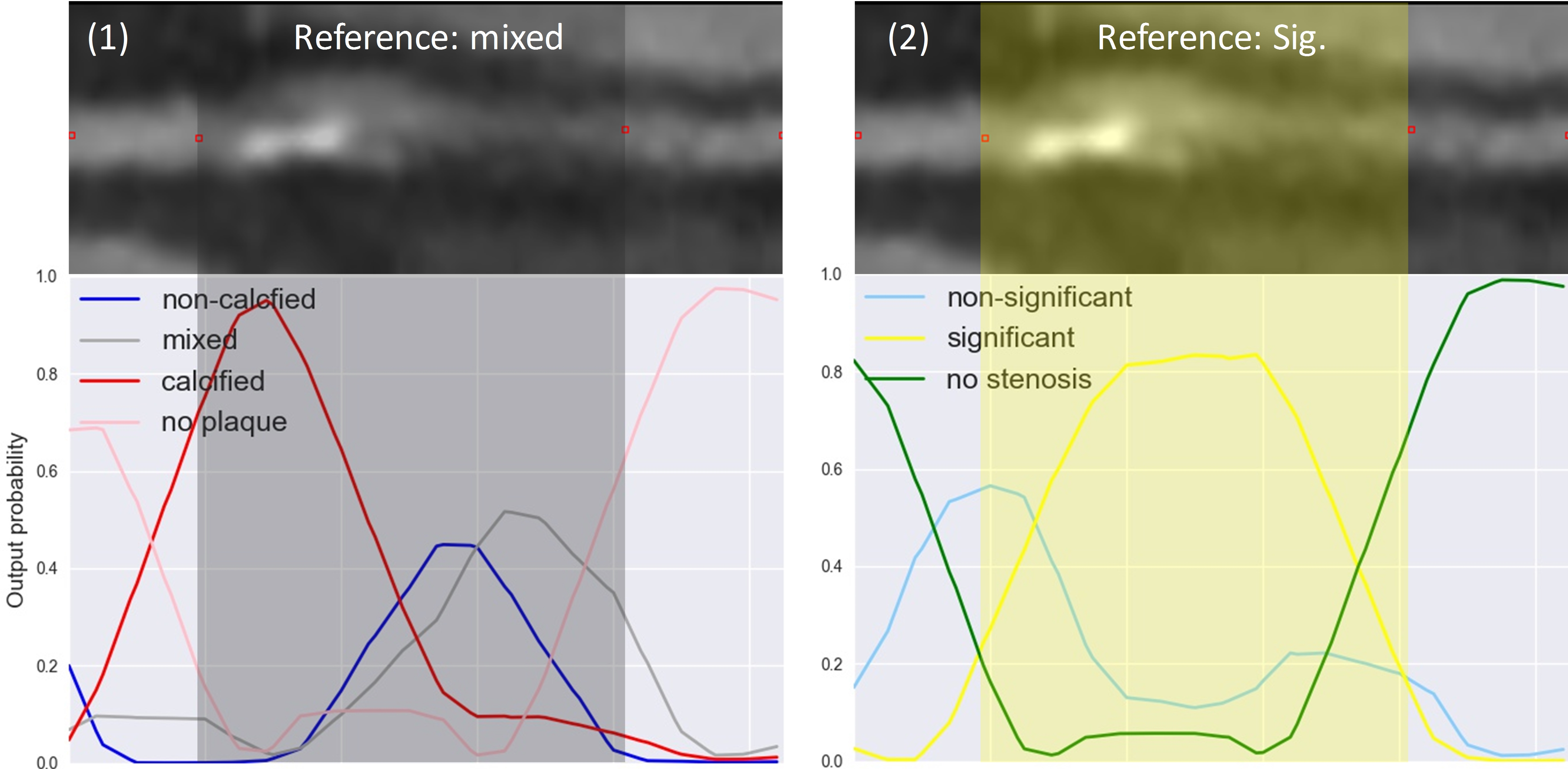}}
		\label{fig:prob_bad}
		\hfil
	
	\subfloat[]{\includegraphics[width=1\linewidth]{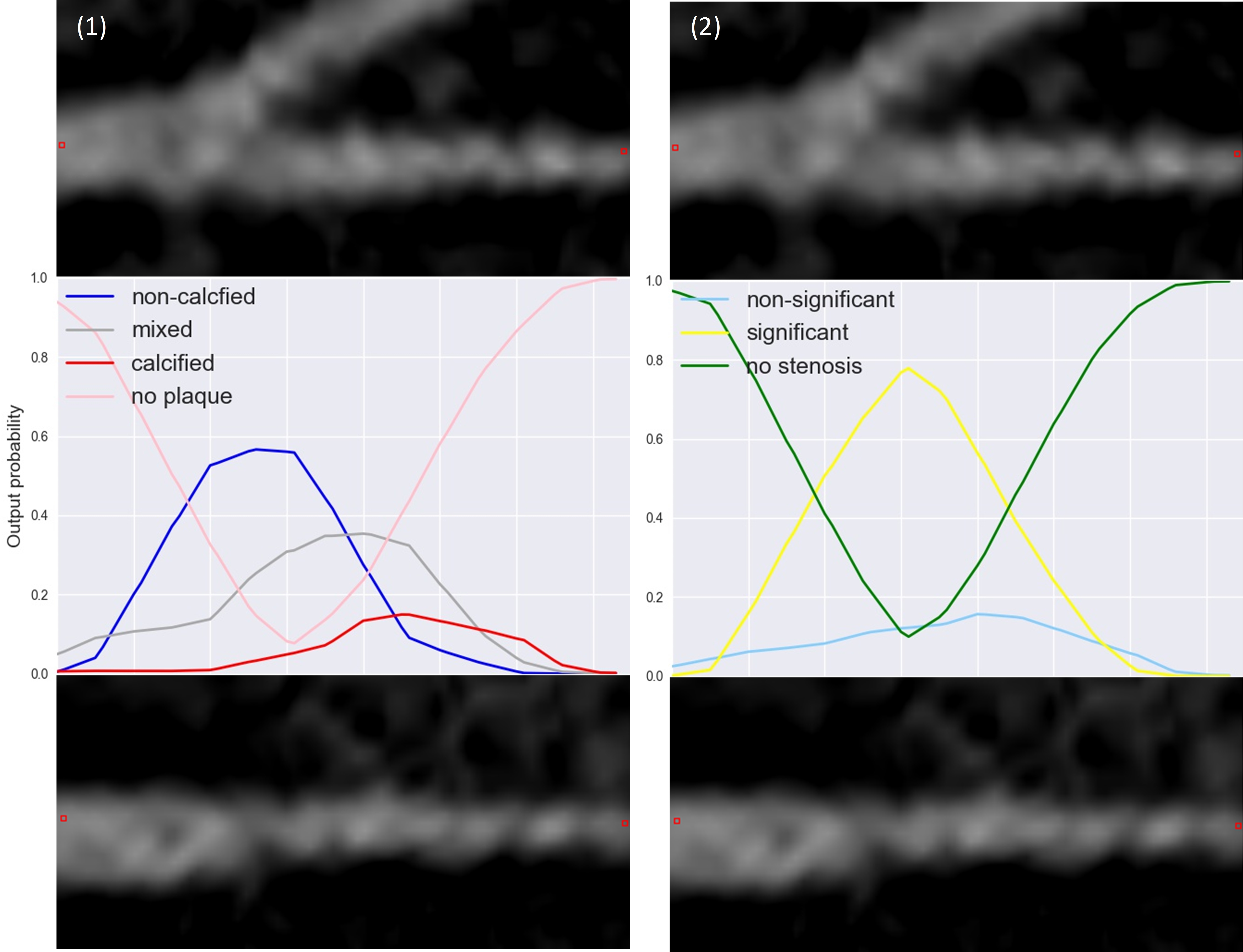}}
		\label{fig:prob_bifu}

	\caption{(a) and (b) are examples of output probabilities for (1) plaque and (2) stenosis classification of a plaque indicated by its manually annotated boundaries and the reference labels (top). Output probabilities for plaque and stenosis classification are shown for each class (bottom). Note that in (a) calcified plaque and non-significant stenosis are correctly detected, while in (b), a mixed plaque with significant stenosis was detected as a calcified plaque with significant stenosis. Non-Sig. = Non-significant stenosis. Sig. = Significant stenosis. Calc. = Calcified plaque. Non-calc. = Non-calcified plaque. (c) An example of output probabilities for (1) plaque and (2) stenosis classification over a bifurcation of an artery in MPR view (top) and a 90$^{\circ}$ rotated MPR view (bottom). Output probabilities for plaque and stenosis classification are shown for each class (middle). Note that as a sudden reduction in the coronary artery lumen diameter occurs distal to bifurcations, in this case, a stenosis was falsely detected.}

	\label{fig:probs}
\end{figure}

\begin{figure*}
	\includegraphics[width=1\linewidth]{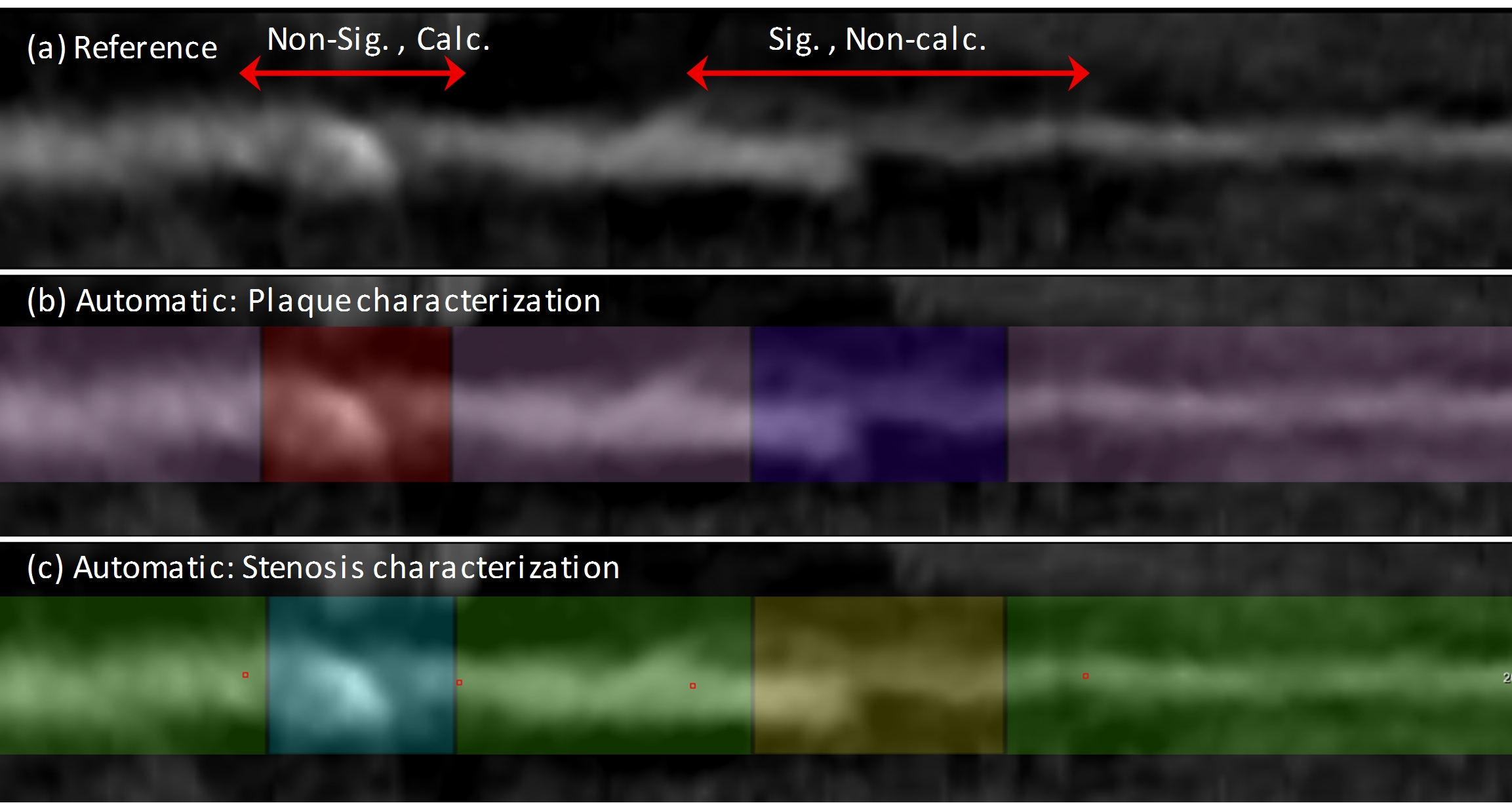}
	\caption{An example of a result for an entire artery. (a) MPR view of an artery with reference standard. Each arrow represents a manually annotated segment, its boundaries and the reference labels. Sig. = Significant stenosis, Non-Sig. = Non-significant stenosis. Calc. = Calcified plaque. Non-calc. = Non-calcified plaque. (b) Automatically predicted plaque labels (no-plaque, calcified, and non-calcified: purple, red, and blue overlays, respectively). (c) Automatically predicted stenosis labels (no-stenosis, non-significant, and significant: green, light blue and yellow, respectively). For illustration purposes only, all overlays were extended to a $25\times25$ voxels around the artery centerline, indicating the analyzed area.}
	\label{fig:prediction_example}
\end{figure*}

\subsection{Plaque detection and characterization}

\add{To assess the performance of the proposed RCNN for plaque detection and characterization (no plaque, non-calcified, mixed, calcified), we first evaluated the performance for each annotated segment. Table~\ref{lesion_type} lists the confusion matrix showing the results.
Next, we evaluated the performance of the proposed method for detection of plaque (plaque of any type vs. no plaque), and compared it with plaque detection and characterization at the segment level. These results are summarized in Table~\ref{perf_summ}. To allow comparison of the automatic analysis with that of an expert, Table~\ref{perf_summ} additionally provides results achieved by the second observer. In brief, for detection of plaque (plaque of any type vs. no plaque), the proposed method achieved a segment-level accuracy of 0.85 and the second observer reached an accuracy of 0.90. For detection and characterization of plaque (no plaque, non-calcified, mixed, calcified), the proposed method and the second observer achieved segment-level accuracies of 0.77 and 0.80 with unweighted $\kappa$ of 0.61 and 0.67, respectively.
Finally, to evaluate whether the performance of the method depends on the analyzed artery, Table~\ref{per_vessel_analysis} lists the performance obtained for the detection and characterization of plaque (no plaque, non-calcified, mixed, calcified) at the segment-level in the different coronary arteries (LAD / LCX / RCA). Plaque was most accurately detected in LCX (0.81) and the least in RCA (0.72). }Examples of automatically predicted classification probabilities in different arteries are shown in Fig.~\ref{fig:probs}. An example of an artery with predicted labels for plaque characterization is shown in Fig.~\ref{fig:prediction_example}(b).

\begin{table}[h!]

	\resizebox{\linewidth}{!}{%
	\centering

	\caption{Confusion matrix showing segment-level results of detection and characterization of plaque by classification into no-plaque, non-calcified plaque, mixed plaque, and calcified plaque. \add{This resulted with an accuracy of 0.77 and an unweighted $\kappa$ of 0.61.}}
	\label{lesion_type}
	\begin{tabular}{ll|cccc}

		\multicolumn{2}{c}{\multirow{2}{*}{\textit{\textbf{Segment-level}}}} & \multicolumn{4}{c}{\textit{Automatic}}  \\
		\multicolumn{2}{c}{} & \multicolumn{1}{l}{No plaque} & Non-calcified & \multicolumn{1}{l}{Mixed} & Calcified \\ \hline
		
		\multirow{4}{*}{\rotatebox[origin=c]{90}{\textit{Reference}}} & No-plaque & \textbf{211} & 8 & 4 & 15 \\
		& Non-calcified & 19 & \textbf{7} & 2 & 0 \\
		& Mixed & 6 & 4 & \textbf{27} & 27 \\
		& Calcified & 11 & 2 & 3 & \textbf{83}
	\end{tabular}}

\end{table}

\begin{table}[h!]
	\centering
	\small
	\caption{Accuracy (Acc.), F1 score and Cohen’s $\kappa$ achieved by the proposed method and by the second observer for plaque and stenosis detection and classification. Results achieved by the second observer are presented between brackets. For analysis of plaque, the performance for detection of plaque (D.) (plaque vs. no plaque), and for detection and characterization of plaque (D.+C.) (no plaque, non-calcified, mixed, calcified) at the segment-level are shown. For analysis of stenosis, the performance for detection of the significant stenosis (D.) (significant stenosis vs. no stenosis or non-significant stenosis), and for detection and classification of the anatomical significance of the stenosis (D.+S.) (no stenosis, non-significant stenosis, significant stenosis), at the segment-, artery- and patients-levels are shown. Note that, for plaque analysis, the unweighted $\kappa$ was computed, while for stenosis analysis, the linearly weighted $\kappa$ was determined.}
\label{perf_summ}
	\begin{tabular}{@{}l|ccccc@{}}
		\multicolumn{1}{c|}{\multirow{2}{*}{\textit{\textbf{\begin{tabular}[c]{@{}c@{}}Plaque\\ Analysis\end{tabular}}}}} & \multicolumn{2}{c}{\textbf{D.}} & \multicolumn{3}{c}{\textbf{D.+C.}} \\
		\multicolumn{1}{c|}{} & Acc. & \multicolumn{1}{c|}{F1} & Acc. & F1 & $\kappa$ \\ \midrule
		\multirow{2}{*}{\textit{Segment-level}} & 0.85 & \multicolumn{1}{c|}{0.85} & 0.77 & 0.61 & 0.61 \\
		& (0.90) & \multicolumn{1}{c|}{(0.90)} & (0.80) & (0.67) & (0.67) \\ \midrule
		\multicolumn{1}{c|}{\multirow{2}{*}{\textit{\textbf{\begin{tabular}[c]{@{}c@{}}Stenosis\\ Analysis\end{tabular}}}}} & \multicolumn{2}{c}{\textbf{D.}} & \multicolumn{3}{c}{\textbf{D.+S.}} \\
		\multicolumn{1}{c|}{} & Acc. & \multicolumn{1}{c|}{F1} & Acc. & F1 & $\kappa$ \\ \midrule
		\multirow{2}{*}{\textit{Segment-level}} & 0.94 & \multicolumn{1}{c|}{0.80} & 0.80 & 0.75 & 0.68 \\
		& (0.91) & \multicolumn{1}{c|}{(0.72)} & (0.83) & (0.72) & (0.70) \\ \midrule
		\multicolumn{1}{c|}{\multirow{2}{*}{\textit{Artery-level}}} & 0.93 & \multicolumn{1}{c|}{0.88} & 0.76 & 0.77 & 0.66 \\
		\multicolumn{1}{c|}{} & (0.92) & \multicolumn{1}{c|}{(0.84)} & (0.80) & (0.78) & (0.72) \\ \midrule
		\multirow{2}{*}{\textit{Patient-level}} & 0.85 & \multicolumn{1}{c|}{0.83} & 0.75 & 0.75 & 0.67 \\
		& (0.83) & \multicolumn{1}{c|}{(0.81)} & (0.75) & (0.77) & (0.70)
	\end{tabular}
\end{table}

\begin{table}[]
	\centering
	\small
	\caption{Accuracy (Acc.), unweighted F1 score and Cohen’s $\kappa$ at the segment-level for plaque and stenosis characterization for the three main coronary arteries (LAD, LCX, RCA). The number of evaluated segments per artery is indicated (n). Note that, for plaque detection and characterization, the unweighted $\kappa$ was computed, while for stenosis detection and determination of the significance of the detected stenosis, linearly weighted $\kappa$ was determined.}
	\label{per_vessel_analysis}
	\begin{tabular}{lcccccc}
		& \multicolumn{3}{c}{\textbf{\textit{Plaque Analysis}}} & \multicolumn{3}{c}{\textbf{\textit{Stenosis Analysis}}} \\
		\multicolumn{1}{c}{} & Acc. & F1 & \multicolumn{1}{c|}{$\kappa$} & Acc. & F1 & $\kappa$ \\ \hline
		\multicolumn{1}{l|}{LAD \small(n=184)} & 0.78 & 0.69 & \multicolumn{1}{c|}{0.65} & 0.81 & 0.79 & 0.72 \\
		\multicolumn{1}{l|}{LCX \small(n=96)} & 0.81 & 0.59 & \multicolumn{1}{c|}{0.62} & 0.80 & 0.65 & 0.60 \\
		\multicolumn{1}{l|}{RCA \small(n=149)} & 0.72 & 0.52 & \multicolumn{1}{c|}{0.53} & 0.79 & 0.68 & 0.62
	\end{tabular}
\end{table}

\subsection{Stenosis detection and characterization}

\add{To assess the performance of the proposed RCNN for detection and determination of the anatomical significance of stenosis, we analyzed the results achieved on the segment-, artery- and patient-levels. The obtained confusion matrices are calculated and shown in Table~\ref{grade}. To get an insight into the ability of the method to classify stenosis, Table~\ref{perf_summ} lists the performance obtained for detection and determination of the anatomical significance of the stenosis (no stenosis, non significant stenosis, significant stenosis) at the segment-, artery- and patient-level. The automatic method achieved accuracy of 0.80, 0.76 and 0,75, and the linearly weighted $\kappa$ of 0.68, 0.66 and 0.67 at the segment-, artery- and patient-level, respectively. For the second observer the accuracies were 0.83, 0.80 and 0.75, and the linearly weighted $\kappa$ were 0.70, 0.72 and 0.70, respectively. Given the importance of the detection of the anatomically significant stenosis, the performance of the automatic method as well as the performance of the second observer for the this task (significant stenosis vs. no stenosis or non-significant stenosis) were evaluated. The automatic method achieved accuracy of 0.94, 0.93 and 0.85 at the segment-, artery- and patient-level, respectively. For the second observer these were 0.91, 0.92 and 0.83, respectively. Table~\ref{perf_summ} also details these results. Finally, to compare the performance in the different coronary arteries, the performance achieved for the detection and determination of the anatomical significance of the stenosis (no stenosis, non-significant stenosis, significant stenosis) at the segment-level in the three main coronary arteries (LAD / LCX / RCA) was evaluated. The analysis showed that similar accuracies were achieved in all three arteries (0.81 for LAD, 0.80 for LCX and 0.79 for RCA). Detailed results are listed in Table~\ref{per_vessel_analysis}.} Examples of automatically predicted classification probabilities in different arteries are shown in Fig.~\ref{fig:probs}. An example of an artery with predicted labels for stenosis characterization is shown in Fig.~\ref{fig:prediction_example}(c).

\begin{table}
	\small
	\resizebox{\linewidth}{!}{%
		
		\caption{Confusion matrices showing segment-, artery-and patient-level results for detection and characterization of the stenosis by classification into no-stenosis, non-significant and significant stenosis. \add{ Accuracies of 0.80, 0.76 and 0.75 were obtained at the segment-, artery- and patient-level, respectively. The linearly weighted $\kappa$ were 0.68, 0.66 and 0.67 at the segment-, artery- and patient-level, respectively.}}		
		
		\label{grade}
		\begin{tabular}{ll|ccc}
			\multicolumn{2}{l}{\multirow{2}{*}{\textit{\textbf{Segment-level}}}} & \multicolumn{3}{c}{\textit{Automatic}} \\
			\multicolumn{2}{l}{} & \multicolumn{1}{l}{No stenosis} & Non-significant & \multicolumn{1}{l}{Significant} \\ \hline 
			
			\multirow{3}{*}{\small\rotatebox[origin=c]{90}{\textit{Reference}}} & No-stenosis & \textbf{211} & 23 & 4 \\
			& Non-significant & 35 & \textbf{112} & 8 \\
			& Significant & 1 & 13 & \textbf{22}\\ & & &
			
	\end{tabular}}
	%\end{table}

	%\begin{table}[]
	
	%	\caption{My caption}
	%\label{vessel_grade}
	
	\resizebox{\linewidth}{!}{%
		\begin{tabular}{llccc}
			\hline
			\multicolumn{2}{l}{\multirow{2}{*}{\textit{\textbf{Artery-level}}}} & \multicolumn{3}{c}{\textit{Automatic}} \\
			\multicolumn{2}{l}{} & \multicolumn{1}{l}{No stenosis} & Non-significant & \multicolumn{1}{l}{Significant} \\ \hline

			\multirow{3}{*}{\small\rotatebox[origin=c]{90}{\textit{Reference}}} & \multicolumn{1}{l|}{No-stenosis} & \textbf{58} & 21 & 3 \\
			& \multicolumn{1}{l|}{Non-significant} & 13 & \textbf{64} & 5 \\
			& \multicolumn{1}{l|}{Significant} & 0 & 5 & \textbf{25}\\ & & &
	\end{tabular}}
	%\end{table}
	
	\resizebox{\linewidth}{!}{%
		\begin{tabular}{llccc}
			\hline
			\multicolumn{2}{l}{\multirow{2}{*}{\textit{\textbf{Patient-level}}}} & \multicolumn{3}{c}{\textit{Automatic}} \\
			\multicolumn{2}{l}{} & \multicolumn{1}{l}{No stenosis} & Non-significant & \multicolumn{1}{l}{Significant} \\ \hline

			\multirow{3}{*}{\small\rotatebox[origin=c]{90}{\textit{Reference}}} & \multicolumn{1}{l|}{No-stenosis} & \textbf{10} & 2 & 1 \\
			& \multicolumn{1}{l|}{Non-significant} & 4 & \textbf{21} & 6 \\
			& \multicolumn{1}{l|}{Significant} & 0 & 3 & \textbf{18}
	\end{tabular}}

\end{table}

\subsection{Impact of the RCNN architecture}\label{cnn_only}
\begin{figure}
	
	\includegraphics[width=0.8\linewidth]{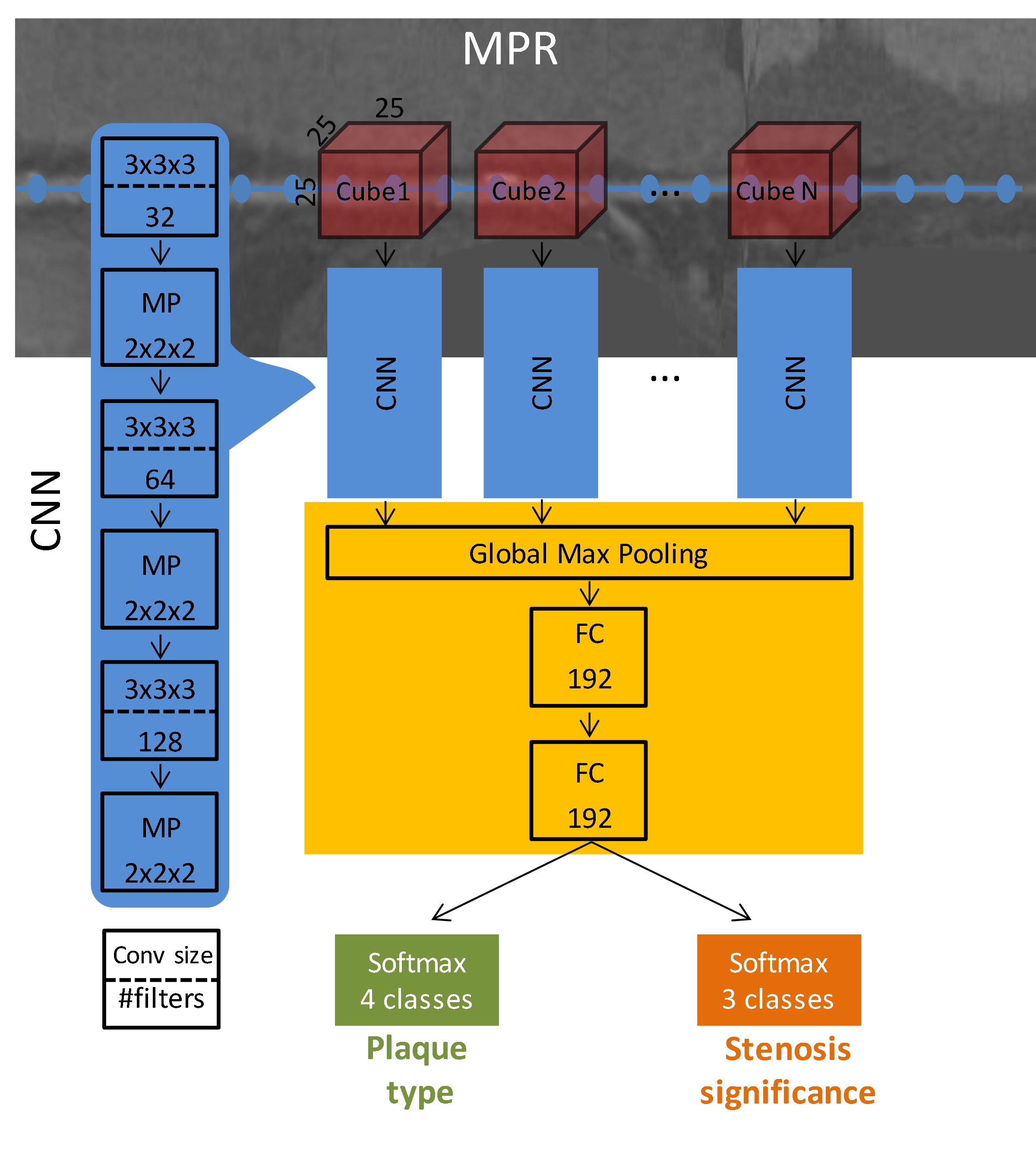}
	
	\caption{
		The input of the network is a sequence of cubes extracted from the MPR along artery centerline. A CNN extracts features out of each $25\times25\times25$ voxels cube, then a global max pooling and two dense \add{(FC - fully connected)} layers process the entire sequence. The output of the \add{dense layers} is fed into two softmax classifiers to simultaneously characterize plaque and stenosis. To match the total number of parameters compared to the proposed network, 196 units in each of dense layers were used. In total, the network had 341,191 parameters, vs. 340,295 parameters in the proposed network.
	}
	\label{fig:cnn_fc}
\end{figure}

To establish the value of the recurrent nature of the proposed network, an additional experiment was performed in which a network with an identical CNN architecture was utilized, while the RNN was replaced by fully connected (FC) layers (Fig.~\ref{fig:cnn_fc}). To analyze different sequence lengths and to aggregate the features extracted by the CNN into one vector, a global max pooling layer was employed after the CNN. This layer was subsequently connected to two FC layers instead of the GRUs. To match the total number of trainable parameters in both architectures, the number of units in each of FC layers was raised from 64 to 192. In total, the network had 341,191 parameters (vs. 340,295 parameters in the RCNN network). To allow a comparison with the proposed RCNN network, this network was trained, validated and tested using the same sets of training, validation and test images. The obtained results are listed in Table~\ref{compare} (\add{second} row).

%\textcolor{red}{MOVE TO TABLE For plaque detection and classification (no-plaque, non-calcified, mixed, calcified) the method achieved accuracy of 0.65, average F1 score of 0.61 and $\kappa$ of 0.49. On a segment-level, for stenosis detection and classification (no-stenosis, non-significant, significant) the method achieved accuracy of 0.75, average F1 score of 0.74 and linearly weighted $\kappa$ of 0.60. On an artery-level, for stenosis detection and classification (no-stenosis, non-significant, significant) the method achieved accuracy of 0.9, average F1 score of 0.61 and linearly weighted $\kappa$ of 0.74.}

\subsection{Single vs. multi-task classification}
Given that plaque and stenoses analyses are related, the classification could be posed as a single multi-class task with seven unique output classes (no plaque, calcified, mixed or non-calcified plaque with non-significant stenosis, calcified, mixed or non-calcified plaque with significant stenosis).
%To evaluate the performance of this merged classification, a network with identical RCNN architecture was utilized, where the two softmax classifiers were replaced by a single softmax classifier with 7 output units.To allow a comparison with the proposed network, this network was trained, validated and tested using the same sets of training, validation and test images. The obtained results are listed in Table~\ref{compare} (\add{first and thirds} rows). 

%R1C4
To evaluate the performance of \add{single task} classification, \add{two additional networks were examined. The first network utilized an identical RCNN architecture with a single softmax classifier with 7 output units. The second used a CNN-only architecture (as in Section \ref{cnn_only}) also with a single softmax classifier with 7 output units.
To allow a comparison with the proposed network, these networks were trained, validated and tested using the same sets of training, validation and test images. The obtained results are listed in Table~\ref{compare} (\add{first and third} rows).}

\begin{table}[h!]

	\centering
	\small
	\caption{Accuracy (Acc.), unweighted F1 score and Cohen’s $\kappa$ at the segment-level for plaque and stenosis characterization using \add{four} different network architectures. \textit{CNN} indicates the network where the recurrent layers were replaced by fully connected layers. \textit{Single-task} denotes the network with single softmax classifier, \add{while \textit{multi-task} denotes the network with two softmax classifiers}. Note that, for plaque detection and characterization, the unweighted $\kappa$ was computed, while for stenosis detection and determination of the significance of the detected stenosis, linearly weighted $\kappa$ was determined.}
	\label{compare}
	\begin{tabular}{lcccccc}
		& \multicolumn{3}{c}{\textbf{\textit{Plaque Analysis}}} & \multicolumn{3}{c}{\textit{\textbf{Stenosis Analysis}}} \\
		\multicolumn{1}{c}{} & Acc. & F1 & \multicolumn{1}{c|}{$\kappa$} & Acc. & F1 & $\kappa$ \\ \hline
		\multicolumn{1}{l|}{\add{CNN single-task}} & \add{0.53} & \add{0.41} & \multicolumn{1}{c|}{\add{0.39}} & \add{0.62} & \add{0.60} & \add{0.46} \\
		\multicolumn{1}{l|}{CNN \add{multi-task}} & 0.63 & 0.58 & \multicolumn{1}{c|}{0.48} & 0.71 & 0.70 & 0.58 \\
		\multicolumn{1}{l|}{\add{RCNN} single-task} & 0.63 & 0.57 & \multicolumn{1}{c|}{0.49} & 0.69 & 0.68 & 0.51 \\
		\multicolumn{1}{l|}{\textbf{\add{RCNN} multi-task}} & \textbf{0.77} & \textbf{0.61} & \multicolumn{1}{c|}{\textbf{0.61}} & \textbf{0.80} & \textbf{0.75} & \textbf{0.68}
	\end{tabular}
\end{table}

%For plaque detection and classification (no-plaque, non-calcified, mixed, calcified) the method achieved accuracy of 0.65, average F1 score of 0.57 and $\kappa$ of 0.49. On a segment-level, for stenosis detection and classification (no-stenosis, non-significant, significant) the method achieved accuracy of 0.71, average F1 score of 0.70 and linearly weighted $\kappa$ of 0.54. On an artery-level, for stenosis detection and classification (no-stenosis, non-significant, significant) the method achieved accuracy of 0.91, average F1 score of 0.62 and linearly weighted $\kappa$ of 0.79.

\subsection{Comparison with previous work}

Most published methods reported segment-level sensitivity and positive predictive value (PPV) for the detection of the anatomically significant stenosis \cite{Kiri13a}. Shahzad et al. \cite{shahzad2012automatic} reported a segment-level sensitivity of 0.50 and a PPV of 0.27 for the detection of the anatomically significant stenosis, while Wang et al. \cite{wang2012vessel} reported a sensitivity of 0.28 and a PPV of 0.23. The proposed network achieved a sensitivity of 0.61 and a PPV of 0.65 for detecting significant stenosis. Nonetheless, comparison with these methods is not trivial as these methods required the artery lumen to be first segmented and then the stenosis was detected. Additionally, different studies reported performance using a different evaluation procedure and different sets of patients than the current work. Therefore, a direct comparison of the results should be used only as an indication.

Moreover, most methods for the automated detection of the calcified coronary artery plaque perform quantification of the plaque \cite{wolterink2016evaluation}. Such methods usually perform voxel-level analysis which our method does not offer. Methods for detection of non-calcified plaque \cite{schepis2009quantification,dey2010automated} perform manual or semi-automatic quantification that requires substantial manual interaction by experts. Therefore, a direct comparison with such methods is not feasible.

\section{Discussion and conclusion}\label{discussion}

A method for automatic detection and characterization of the coronary artery plaque type, as well as detection and characterization of the anatomical significance of the coronary artery stenosis was presented. The method employs an RCNN that analyzes an MPR view of a coronary artery extracted from a CCTA scan using the coronary artery centerline. The RCNN utilizes a 3D CNN that computes image features from 3D volumes extracted along the coronary artery centerline. Subsequently, an RNN analyzes the computed image features to perform both classification tasks. Unlike most previous methods that detect and characterize coronary artery plaque and stenosis relying on the coronary artery lumen segmentation \cite{Kiri13a}, the proposed method requires only the coronary artery centerline as an input along with the CCTA image.

The presented results reveal that detection and characterization of coronary artery plaque can be performed accurately, but with moderate reliability (Table~\ref{lesion_type} and Table~\ref{perf_summ}). For both plaque detection and plaque characterization, the second observer only achieved slightly better performance than the proposed method (Table~\ref{perf_summ}) indicating the complexity of the task. 
Nevertheless, the method was accurate in discriminating segments with plaque from those without any plaque. This is clinically important as absence of plaque does not lead to treatment. Moreover, further analysis of plaque characterization revealed that differentiation of the mixed plaque from the calcified and non-calcified plaque remains challenging. This is not surprising given that the mixed plaque contains both calcified and non-calcified components and that the distinction between mixed and calcified plaque, as well as between mixed and non-calcified plaque, is not clearly defined as illustrated in Fig.~\ref{fig:prob_bad}. To address this, the automatic method could perform detection of calcified and non-calcified components only, and the obtained results could be merged into calcified, non-calcified and mixed plaque based on their spatial distribution. In addition, it would be interesting to segment plaque on a voxel-level, but obtaining voxel-wise reference is extremely labor intensive and requires a very experienced expert. Furthermore, analysis of the results per coronary artery (Table~\ref{per_vessel_analysis}) reveals that plaque characterization achieves a slightly lower performance in segments located in the RCA than in LAD or LCX. This might be caused by the more prominent cardiac motion artifacts in the RCA compared to the other two arteries \cite{hong2001ecg}.

Anatomically significant stenosis could potentially lead to myocardial ischemia (i.e. functionally significant stenosis), and clinical guidelines suggest that different grades of stenosis in the coronary artery should be managed differently \cite{cury2016cad}. Our experiments demonstrate that the proposed method is able to detect and determine the anatomical significance of coronary artery stenosis accurately with excellent reliability (Table~\ref{perf_summ} and Table~\ref{grade}). For detection and characterization of the anatomical significance of a stenosis, the proposed method achieved a performance approaching the level of the second observer (Table~\ref{perf_summ}). Moreover, we investigated whether the method is able to identify patients with anatomically significant stenosis. This is especially important as patients having such stenosis are usually referred to further functional testing and to invasive coronary angiography to measure fractional flow reserve (FFR). FFR determines the functional significance of the stenoses, and hence, establishes the patient's treatment strategy. Our results reveal that patients with anatomically significant stenosis on CCTA are detected with high accuracy (Table~\ref{perf_summ}). However, one patient without stenosis was identified as having a significant stenosis (Table~\ref{grade}). An examination of the CCTA image of this patient revealed that although no artifacts were present, a coronary artery bifurcation was mistakenly detected as significant stenosis in one of the patients' arteries.
Moreover, in this work, only two distinct degrees of stenosis were differentiated; below and above 50\% of luminal narrowing. Future work may investigate automatic classification of additional clinically relevant stenotic grades, e.g. $<$25\% or $>$70\%, or automatic estimation of stenosis degree, i.e. percentage of luminal narrowing. Nevertheless, for both tasks, a larger training set of patients with manual annotations would be required.

The contribution of the recurrent nature of the proposed network was evaluated. The results show clear advantage of the proposed recurrent architecture over the network containing no recurrent units (Table~\ref{compare}). This is in agreement with our assumption that a sequential analysis of several small volumes along the coronary artery is needed to aggregate the knowledge of the entire analyzed region rather than just locally (e.g. a single volume). A similar concept was reported by Ng et al. \cite{ng2015beyond}, where incorporating information across video frames enabled better video classification.
Possibly, given a sufficiently large and diverse data set, a \addminor{deeper} CNN-only (e.g. 3D U-Net \cite{cciccek20163d}), analyzing a large single volume along the artery, could be employed to perform the presented analyses. However, obtaining such a large data set remains highly challenging, \addminor{and employing deep networks, typically used for analysis of natural images (e.g. ResNet \cite{he2015deep}), is likely not beneficial due to scarcity of the manually labeled training data.}

Unlike most methods for coronary artery plaque and stenosis classification that depend on the extracted artery centerline followed by arterial lumen segmentation \cite{Kiri13a}, the proposed method relies only on the extracted artery centerline. Arterial lumen segmentation is far from trivial task, which occasionally requires substantial manual interaction, especially in diseased population with heavily calcified arteries. We have here prevented potential error propagation by omitting this step. To extract coronary artery centerlines, we have employed our previously designed method for artery centerline extraction \cite{wolterink2018coronary}. However, any other manual, semi-automatic or automatic method could be employed instead.  %R1M1
\add{ Although employing the extracted centerlines and the subsequent analysis of the MPR images simplify plaque and stenosis classification, small errors in centerline extraction might lead to errors and therefore negatively affect the overall performance. To mitigate impact of such errors, we trained the RCNN with augmented centerlines, simulating small inaccuracies (Fig.\ref{learning_curve}b). Future work might investigate direct classification of plaque and stenosis from acquired CCTA images, omitting the intermediate centerline extraction.}  % Although the employed artery centerline extraction method is near real time, where a single artery is tracked in $< 1$ second, it still requires manual seed placement. .

In this work, we have treated plaque and stenosis characterization as two different tasks, that were performed jointly. Although this halved the time of inference (1.8 seconds per artery on average), it has a limitation. Given that the parameters of the two softmax classifiers in the network differ, a physiologically impossible scenario, where a plaque is not detected while a stenosis is detected, can occur. Although in our experiments this was the case only in less than 1.5\% of the cases%251/16428),
, future work could address this either by modifying the network architecture preventing such scenario, or by applying a high penalty for such cases in the loss function of the network. A single task network would prevent such a scenario. Nevertheless, \add{experiments}, comparing the proposed multi-task \add{approach} against \add{networks} with \add{single softmax} classifier each, demonstrate superior performance of the multi-task approach (Table~\ref{compare}, \add{first and third rows vs. second and fourth rows}). The limited number of training samples may have prevented the single-task networks from generalizing well even with a relatively small network. \add{Moreover, although the architecture of the RCNN was determined in preliminary experiments using the training set, a systematic extensive grid-search or other hyper-parameter optimization methods were not performed. Addressing this might further improve the results.} %In future work, size of the dataset may be increased by obtaining manual annotations in more images as well as by utilizing additional data augmentation methods.
 
In the manual annotations defining the reference standard, a single label was assigned to a whole segment of the coronary artery containing plaque. Separating these segments to their local components might lead to different labels. Consequently, the identified start- and end-points of the automatically detected plaque and stenosis are not in full agreement with the reference annotations (see Fig.~\ref{fig:prediction_example} and Fig.~\ref{fig:prob_bad}). Furthermore, coronary artery bifurcations were not manually annotated and the network was not trained to detect these as a separate class. As a sudden reduction in the coronary artery lumen diameter occurs distal to bifurcations \cite{antiga2004robust}, a stenosis might be \add{mistakenly} detected (\add{see example illustrated in Fig.~\ref{fig:prob_bifu}}). Future work might address these limitations by modifying the reference standard so that each voxel in the arterial wall or each cross-section of the arterial lumen is annotated and the coronary artery bifurcations are indicated. Besides bifurcations, imaging artefacts caused by routinely used step-and-shoot protocol could affect performance of the method. Nonetheless, a qualitative evaluation of the results revealed that arterial segments containing such image artifacts were not incorrectly detected as having plaque or stenosis. Note that these regions were not included in the manual annotations either.
%R2C1
\add{Finally, the current study employed clinically obtained CCTA scans from a single vendor. Modern scanners applying new techniques (e.g. high-pitch spiral, 320-detector row) may improve image quality and potentially enable further increase in the performance of the here proposed method. Moreover, further studies are needed to evaluate the proposed method using a larger set of scans from different vendors and medical centers.}

To conclude, this study presented an algorithm, based on a recurrent convolutional neural network, for automatic detection and characterization of coronary artery plaque, as well as detection and characterization of the anatomical significance of coronary artery stenosis. To the best of our knowledge, we are the first to propose an automatic method for both plaque and stenosis characterizations. This may enable automated triage of patients to those without coronary plaque, and those with coronary plaque and stenosis in need for further cardiovascular investigation. 

\small
\section*{Acknowledgments}
	
This study was financially supported by the project FSCAD, funded by the Netherlands Organization for Health Research and Development (ZonMw) in the framework of the research programme IMDI (Innovative Medical Devices Initiative); project 104003009.

	%\section*{References}
	%\medskip

	\small
% Generated by IEEEtran.bst, version: 1.14 (2015/08/26)

\end{document}